\documentclass[12pt]{article}
\pdfoutput=1
\usepackage{graphicx,pict2e,latexsym,xcolor,amsmath,amssymb}
\usepackage[square,numbers]{natbib}
\newtheorem{theorem}{Theorem}

\newtheorem{example}{Example}

\begin{document}
\bibliographystyle{plainnat}
\pagestyle{plain}

\title{\Large \bf Out-of-Sample Embedding with Proximity Data: 
Projection versus Restricted Reconstruction}

\author{Michael W. Trosset\thanks{Department of Statistics, Indiana University.  
E-mail: {\tt mtrosset@iu.edu}} 
\and
Kaiyi Tan\thanks{Department of Statistics, Indiana University.  
E-mail: {\tt ddhp0704@gmail.com}}
\and
Minh Tang\thanks{Department of Statistics, North Carolina State University.  
E-mail: {\tt mtang8@ncsu.edu}} 
\and
Carey E. Priebe\thanks{Department of Applied Mathematics \& Statistics, Johns Hopkins University. 
E-mail: {\tt cep@jhu.edu}}
}

\date{\today}
%\date{October 3, 2011}

\maketitle

%\newpage

\begin{abstract}
  
\end{abstract}

The problem of using proximity (similarity or dissimilarity) data for the purpose of ``adding a point to a vector diagram'' was first studied by J.~C. Gower in 1968.  Since then, a number of methods---mostly kernel methods---have been proposed for solving what has come to be called the problem of {\em out-of-sample embedding}.  We survey the various kernel methods that we have encountered and show that each can be derived from one or the other of two competing strategies: {\em projection}\/ or {\em restricted reconstruction}.  Projection can be analogized to a well-known formula for adding a point to a principal component analysis.  Restricted reconstruction poses a different challenge: how to best approximate redoing the entire multivariate analysis while holding fixed the vector diagram that was previously obtained.  This strategy results in a nonlinear optimization problem that can be simplified to a unidimensional search.  Various circumstances may warrant either projection or restricted reconstruction.


\newpage

\tableofcontents

\newpage

%=====

\section{Introduction}
\label{intro}

The embedding method that is now called {\em classical multidimensional scaling}\/ (CMDS) was proposed by Torgerson \cite{torgerson:1952}.
A seminal paper by Gower \cite{gower:1966} characterized CMDS as a Q technique, i.e., a technique that depends on a set of points through their pairwise interpoint distances.  As such distances can be computed from pairwise inner products, CMDS is, in the parlance of our times, a {\em kernel method}.  Subsequently, in
a pioneering example of what is now termed {\em out-of-sample embedding}, Gower \cite{gower:1968} considered the problem of ``adding a point to vector diagrams in multivariate analysis'' using information about pairwise interpoint distances. 

Since Gower's contributions in the 1960s, various researchers have considered out-of-sample embedding problems with both similarity and dissimilarity data.  The present investigation was motivated by our efforts to understand and reconcile the differences between the methods proposed in \cite{bengio&etal:2003tr,bengio&etal:2003} 
and in \cite{mwt:out}.  Briefly, the method in \cite{mwt:out} is derived by approximating entries in a certain matrix, resulting in a polynomial objective function that contains a quadratic and a quartic term.  The quadratic term results from certain off-diagonal entries, whereas the quartic term results from a diagonal entry.  Discarding the quartic term results in a linear least squares problem, the solution of which is the embedding formula proposed by Bengio et al.\ 
\cite{bengio&etal:2003tr,bengio&etal:2003}, who explicitly rejected consideration of the diagonal entry and (therefore) the possibility of minimizing the entire quartic polynomial.  In contrast, the authors of \cite{mwt:out} long believed the quartic polynomial to be the appropriate objective function and viewed its quadratic term as a simplifying approximation of it.  Only recently have we come to understand that each solution can be derived from (different) first principles.

By a {\em vector diagram}, Gower \cite{gower:1968} meant a Cartesian representation of $n$ points in $d$-dimensional Euclidean space.
Key to our development is the distinction between the representation space and the configuration of $n$ points within it.  Out-of-sample embedding fixes a configuration of points in the sense that it fixes the corresponding set of pairwise interpoint distances.  It may or may not fix the space in which this configuration is represented.  As it turns out, both \cite{gower:1968} and 
\cite{bengio&etal:2003tr,bengio&etal:2003} 
fix the representation space, whereas \cite{mwt:out} does not.

We elaborate on the distinction between representation space and configuration of points in Section~\ref{vs}.  We identify two fundamental embedding strategies and illustrate their consequences in the familiar case of classical principal component analysis (PCA).
Following previous exposition in an unpublished technical report \cite{tang&trosset:out2},
we demonstrate in Section~\ref{project} that most well-known methods for out-of-sample embedding, including \cite{gower:1968} and \cite{bengio&etal:2003}, follow from the first strategy.
In Section~\ref{rr} we implement the second strategy, culminating in the method proposed in \cite{mwt:out}.  The second set of methods present computational challenges, which we discuss in Section~\ref{compute}.  Section~\ref{disc} discusses how a user might choose a suitable strategy.

\section{Two Fundamental Strategies}
\label{vs}

\begin{figure}[tb]
\begin{center}
\fbox{ 
\begin{minipage}{5in}
\vspace{1em}
\begin{enumerate}

\item Projection.  Construct an ambient space that contains the original representation space.  Position the new point in the ambient space, then project it into the representation space.

\item Restricted Reconstruction.  Suppose that the original configuration was obtained by optimizing a function of $n$ points.  Minimize the analogous function of $n+1$ points, subject to an additional constraint that fixes the interpoint distances between the original $n$ points, to obtain a new configuration of $n+1$ points.

\end{enumerate}
\vspace{1em}
\end{minipage}
}
\end{center}
\caption{Two fundamental strategies for out-of-sample embedding.  Both strategies fix the original configuration of points.  The projection strategy also fixes the representation space.  Unlike the projection strategy, the restricted reconstruction strategy depends on the method used to construct the original configuration.}
\label{fig:strategies}
\end{figure}

Figure~\ref{fig:strategies} provides brief summaries of two strategies for out-of-sample embedding.  To appreciate the nuances of these strategies, suppose that we have used PCA to construct a $d$-dimensional representation of feature vectors $\xi_1,\ldots,\xi_n \in \Re^q$.
We subsequently observe an additional feature vector $\eta \in \Re^q$.
How might we construct a $d$-dimensional representation of $\xi_1,\ldots,\xi_n,\eta$?  Of course, we might simply set $\xi_{n+1}=\eta$ and apply PCA to $\xi_1,\ldots,\xi_{n+1}$.  We refer to this practice as {\em reconstruction}.  But perhaps we have already used the original representation for some purpose and wish to preserve some or all of its features.  This is the consideration that motivates the problem of {\em out-of-sample embedding}.

PCA may be characterized as the method of finding the $d$-dimensional affine linear subspace (hyperplane) ${\mathcal H} \subset \Re^q$ that minimizes the sum of the squared Euclidean distances of the $\xi_i$ from ${\mathcal H}$.  (Other characterizations are possible; another turns out to be more useful when embedding proximity data.)  PCA adopts a particular coordinate system for ${\mathcal H}$, thereby identifying it with $\Re^d$.  This is the representation space constructed by PCA, and the $x_1,\ldots,x_n \in \Re^d$ that correspond to $\xi_1,\ldots,\xi_n \in \Re^q$ (often called the PC scores) constitute a configuration of points in representation space.  Reconstruction, i.e., application of PCA to $\xi_1,\ldots,\xi_{n+1} \in \Re^q$, results in a new approximating hyperplane, hence a new representation space, and a new configuration of points.

Suppose, however, that we wish to add a point $y \in \Re^d$ that corresponds to $\eta \in \Re^q$ while holding fixed both the representation space and the configuration of points within it.  Fixing the representation space means fixing ${\mathcal H}$, so the natural way to represent $\eta$ is to project $\eta$ into ${\mathcal H}$, obtaining $\hat{\eta}$, then represent $\hat{\eta}$ in PC coordinates, obtaining $\hat{y} \in \Re^d$.  This construction illustrates the projection strategy for out-of-sample embedding.

Alternatively, suppose that we are prepared to adopt a new representation space but wish to hold fixed the representation of $\xi_1,\ldots,\xi_n$, e.g., by fixing the pairwise interpoint distances of $x_1,\ldots,x_n \in \Re^d$.  This is reconstruction subject to a constraint, hence the phrase {\em restricted reconstruction}.  The restricted reconstruction strategy for out-of-sample embedding answers the following question: how does PCA perform on $\xi_1,\ldots,\xi_{n+1} \in \Re^q$ if the representation of $\xi_1,\ldots,\xi_n$ is fixed?  The answer depends on how one chooses to characterize PCA.

A simple example illustrates the distinction between projection and restricted reconstruction.

\begin{example}
Consider $\xi_1,\xi_2 \in \Re^2$ with $\xi_1=(-1,0)$ and $\xi_2=(1,0)$.  The first PC axis is evidently the first (horizontal) axis of $\Re^2$, with PC scores $x_1=-1$ and $x_2=1$.
We then observe $\eta=(0,9)$ and seek to represent $\xi_1,\xi_2,\eta$ in a single dimension.

Reconstruction performs PCA on $\xi_1,\xi_2,\xi_3=\eta$.  The first PC axis is evidently the second (vertical) axis of $\Re^2$, with PC scores $x_1=-3,x_2=-3,x_3=6$.

To retain both the representation space and the configuration of points within it, project $\eta$ into the horizontal axis of $\Re^2$, obtaining $\hat{y}=0$ and $\hat{\eta}=(0,0)$.  Notice that the configuration of points obtained by projection differs substantially from the configuration of points obtained by reconstruction.

How to implement restricted reconstruction is less clear.  The general approach described in Figure~\ref{fig:strategies} requires us to identify the optimization problem solved by reconstruction and add to it constraints that fix the original configuration.  For PCA, there is more than one way to proceed.

An obvious possibility is to characterize PCA as finding the $d$-dimensional hyperplane that best approximates a set of feature vectors.
Let
\[
M = \left[ \begin{array}{rr}
-1 & -3 \\ 1 & -3 \\ 0 & 6
\end{array} \right],
\]
let
\[
V(\theta) = \left[ \begin{array}{rr}
\cos \theta & - \sin \theta \\
\sin \theta & \cos \theta
\end{array} \right],
\]
and let
\[
M(y) = \left[ \begin{array}{rr}
-1-y/2 & 0\\ 1-y/2 & 0 \\ y & 0
\end{array} \right].
\]
Notice that we have translated the original $(\xi_1,\xi_2)$ in order to center the new configuration.  After determining the new configuration, one can reverse the translation to recover the original $(\xi_1,\xi_2)$.

We seek $\theta,y \in \Re$ such that $(\theta,y)$ minimizes
\[
f(\theta,y) = \left\| M V(\theta) - M(y) \right\|_F^2
= 1.5y^2-18y \sin \theta - 4 \cos \theta +58,
\]
where $\| \cdot \|_F$ denotes Frobenius norm.
Taking derivatives,
\begin{eqnarray*}
\nabla f(\theta,y) = \left[ \begin{array}{c}
-18y \cos \theta + 4 \sin \theta \\
3y-18 \sin \theta
\end{array} \right]
 & \mbox{and} &
\nabla^2 f(\theta,y) = \left[ \begin{array}{cc}
18y \sin \theta + 4 \cos \theta & -18 \cos \theta \\
-18 \cos \theta & 3
\end{array} \right].
\end{eqnarray*}
Setting $\nabla f(\theta,y)=\vec{0}$, we obtain one stationary point at $(0,0)$ and other stationary points $(\theta,y)$ that satisfy the equations $\cos \theta =1/27$ and $y=6 \sin \theta$.

The stationary point $(0,0)$ corresponds to the configuration produced by projection.  Notice that it is a saddle point for restricted reconstruction, not a local minimizer.  The other stationary points are local minimizers of $f$, which attains the same minimal value at each of them.  Hence, each of these points is a global minimizer.  Choosing $y = 6 (1-1/27^2)^{1/2}$ gives the configuration
\[
M_1 \doteq \left[ \begin{array}{rr}
-3.998 & 0 \\
-1.998 & 0 \\
5.996 & 0
\end{array} \right],
\]
whereas choosing $y = -6 (1-1/27^2)^{1/2}$ gives the configuration
\[
M_2 \doteq \left[ \begin{array}{rr}
1.998 & 0 \\
3.998 & 0 \\
-5.996 & 0
\end{array} \right].
\]
In the first case, adding $y/2 \doteq 2.998$ to the first coordinate recovers the original $(\xi_1,\xi_2)$ and results in the out-of-sample embedding $x_1 = -1$, $x_2 = 1$, and $y_* \doteq 8.994$.
In the second case, adding $y/2 \doteq -2.998$ to the first coordinate recovers the original $(\xi_1,\xi_2)$ and results in the out-of-sample embedding $x_1 = -1$, $x_2 = 1$, and $y_* \doteq -8.994$.
Notice that either of these configurations more closely resembles the configuration obtained by reconstruction than does the configuration obtained by projection.
\label{ex:3ptOut}
\end{example}

Example~\ref{ex:3ptOut} illustrates a fundamental distinction between projection and restricted reconstruction.  Projection ignores any differences between $\eta$ and $\xi_1,\ldots,\xi_n$ that are orthogonal to the hyperplane that approximates $\xi_1,\ldots,\xi_n$, whereas restricted reconstruction searches for a potentially different hyperplane that approximates such differences.  Example~\ref{ex:3ptOut} is dramatic: the differences between $\eta$ and $\xi_1,\ldots,\xi_n$ are profound and the two hyperplanes are orthogonal.  Absent context, however, we do not claim that either method is superior.  In some applications one may want to ignore orthogonal differences; in other applications, one may want to incorporate such differences into a new representation space.  If the orthogonal distances are small, then projection and restricted representation should yield comparable solutions.

The remainder of this section elaborates on the projection method.  We defer further discussion of the restricted reconstruction method to Section~\ref{rr}, in which necessary tools are developed.

For $\xi_1,\ldots,\xi_n \in \Re^q$, let $\tilde{\xi}_i = \xi_i-\bar{\xi}$,
let $\tilde{M}$ denote the corresponding $n \times q$ centered data matrix,
and let
\[
\tilde{M} = U \left[ \begin{array}{c|c}
\Sigma & 0 \\ \hline 0 & 0 \end{array} \right] V^t
\]
denote the singular value decomposition of $\tilde{M}$.
The symmetric $n \times n$ matrix of centered pairwise inner products is then
\[
B = \tilde{M} \tilde{M}^t = U \left[ \begin{array}{c|c}
\Sigma & 0 \\ \hline 0 & 0 \end{array} \right] V^t V \left[ \begin{array}{c|c}
\Sigma & 0 \\ \hline 0 & 0 \end{array} \right] U^t = U \left[ \begin{array}{c|c}
\Sigma^2 & 0 \\ \hline 0 & 0 \end{array} \right] U^t,
\]
and the $d$-dimensional PCA representation of $\xi_i$ is row $i$ of the $n \times d$ matrix $U_d \Sigma_d$, 
where $\Sigma_d$ is the $d \times d$ diagonal matrix that contains the $d$ largest singular values of $\tilde{M}$ and $U_d$ is the $n \times d$ matrix that contains the corresponding left singular vectors.  

Assuming that $\sigma_d>0$, write 
\begin{eqnarray*}
B U_d \Sigma_d^{-1} & = &
U \left[ \begin{array}{c|c} \Sigma^2 & 0 \\ \hline 0 & 0 \end{array} \right]
U^t U_d \Sigma_d^{-1} \\ & = &
\left[ \begin{array}{c|c} U_d & \cdot \end{array} \right]
\left[ \begin{array}{c|c} \Sigma^2 & 0 \\ \hline 0 & 0 \end{array} \right]
\left[ \begin{array}{c} U_d^t \\ \hline \cdot \end{array} \right]
U_d \Sigma_d^{-1} \\ & = &
\left[ \begin{array}{c|c} U_d & \cdot \end{array} \right]
\left[ \begin{array}{c|c} \Sigma^2 & 0 \\ \hline 0 & 0 \end{array} \right]
\left[ \begin{array}{c} I_d \\ \hline 0 \end{array} \right]
\Sigma_d^{-1} \\ & = &
\left[ \begin{array}{c|c} U_d & \cdot \end{array} \right]
\left[ \begin{array}{c} \Sigma_d^2 \\ \hline 0 \end{array} \right]
\Sigma_d^{-1} \\ & = &
\left[ \begin{array}{c|c} U_d & \cdot \end{array} \right]
\left[ \begin{array}{c} \Sigma_d \\ \hline 0 \end{array} \right] \\ & = &
U_d \Sigma_d = 
\left[ \begin{array}{c} x_1^t \\ \vdots \\ x_n^t \end{array} \right].
\end{eqnarray*}
Letting $b_i$ denote column $i$ of the symmetric matrix $B$, 
the $d$-dimensional PCA representation of $\xi_i$ can then be written as
\begin{eqnarray*}
x_i^t = b_i^t U_d \Sigma_d^{-1} & \mbox{ or } &
x_i = \Sigma_d^{-1} U_d^t b_i.
\end{eqnarray*}

We wish to insert another $\eta \in \Re^q$ into the $d$-dimensional PCA representation of $\xi_1,\ldots,\xi_n$.
Let ${\mathcal L}$ denote the linear subspace of $\Re^q$ spanned by $u_1,\ldots,u_d$.
Each $x_i$ is computed by (1) projecting $\tilde{\xi}_i$ into ${\mathcal L}$,
then (2) rotating ${\mathcal L}$ to a representation in which the first $d$ coordinates correspond to $u_1,\ldots,u_d$ and the remaining $n-d$ coordinates vanish, and finally (3) discarding the superfluous coordinates.
We seek to do likewise for $\eta$, obtaining $y \in \Re^d$.

Write $\tilde{\eta} = \eta-\bar{\xi}$ as
\[
\tilde{\eta} = \sum_{i=1}^n \alpha_i \tilde{\xi}_i+w,
\]
where $w$ is orthogonal to the span of $\tilde{\xi}_1,\ldots,\tilde{\xi}_n$ and therefore to ${\mathcal L}$.  Letting $\pi_{\mathcal L}$ denote projection into ${\mathcal L}$, it follows that
\[
\hat{y} = \pi_{\mathcal L} \left( \tilde{\eta} \right) =
\sum_{i=1}^n \alpha_i \pi_{\mathcal L} \left( \tilde{\xi}_i \right) + 
\pi_{\mathcal L} (w) =
\sum_{i=1}^n \alpha_i \pi_{\mathcal L} \left( \tilde{\xi}_i \right)
 = \sum_{i=1}^n \alpha_i x_i.
\]
Rotating ${\mathcal L}$ to coordinate axes $u_1,\ldots,u_d$ will not affect the representation of $\pi_{\mathcal L} \left( \tilde{\eta} \right)$ as a linear combination of the $\pi_{\mathcal L} \left( \tilde{\xi}_i \right)$,
nor will discarding superfluous coordinates; hence,
\begin{eqnarray} \nonumber
\hat{y} & = & \sum_{i=1}^n \alpha_i x_i = 
\sum_{i=1}^n \alpha_i \Sigma_d^{-1} U_d^t b_i =
\Sigma_d^{-1} U_d^t \sum_{i=1}^n \alpha_i b_i \\ \nonumber
 & = & \Sigma_d^{-1} U_d^t \sum_{i=1}^n \alpha_i 
\left[ \begin{array}{c} \left\langle \tilde{\xi}_1,\tilde{\xi}_i \right\rangle \\ 
\vdots \\ \left\langle \tilde{\xi}_n,\tilde{\xi}_i \right\rangle \end{array} \right] =
\Sigma_d^{-1} U_d^t \left[ \begin{array}{c} \left\langle \tilde{\xi}_1,
\sum_{i=1}^n \alpha_i \tilde{\xi}_i \right\rangle \\ 
\vdots \\ \left\langle \tilde{\xi}_n,
\sum_{i=1}^n \alpha_i \tilde{\xi}_i \right\rangle \end{array} \right] \\
 & = & \Sigma_d^{-1} U_d^t \left[ \begin{array}{c} \left\langle \tilde{\xi}_1,
\sum_{i=1}^n \alpha_i \tilde{\xi}_i +w \right\rangle \\ 
\vdots \\ \left\langle \tilde{\xi}_n,
\sum_{i=1}^n \alpha_i \tilde{\xi}_i +w \right\rangle \end{array} \right] =
\Sigma_d^{-1} U_d^t \left[ \begin{array}{c} \left\langle \tilde{\xi}_1,
\tilde{\eta} \right\rangle \\ 
\vdots \\ \left\langle \tilde{\xi}_n,
\tilde{\eta} \right\rangle \end{array} \right] =
\Sigma_d^{-1} U_d^t b
\label{eq:pca.x}
\end{eqnarray}
is an out-of-sample projection formula for PCA that relies on centered
Euclidean inner products.  We will exploit this formula in Sections \ref{project} and \ref{rr}.

Finally, there is an equivalent projection formula that relies on squared Euclidean distances.  Write
\[
\left\| \tilde{\xi}_i -\tilde{\eta} \right\|^2 = 
\left\| \tilde{\xi}_i \right\|^2 -
2 \left\langle \tilde{\xi}_i,\tilde{\eta} \right\rangle + \| \tilde{\eta} \|^2
\]
and solve for $\langle \tilde{\xi}_i,\tilde{\eta} \rangle$ to obtain element $i$ of $b$.
Substituting this expression into (\ref{eq:pca.x}) then yields out-of-sample embedding formula (10) in \cite{gower:1968}.

\section{Projection with Proximity Data}
\label{project}

We now turn to the case of proximity data.  Instead of assuming that $\xi_1,\ldots,\xi_n$ are observed feature vectors that reside in $\Re^q$, we suppose that $\xi_1,\ldots,\xi_n$ are unobserved objects that reside in a space $\Xi$.  What we know about these objects are their pairwise proximities, either similarities or dissimilarities.
Formally, we say that a symmetric matrix $\Gamma = [ \gamma_{ij} ]$ is a {\em similarity matrix}\/ if and only if each $0 \leq \gamma_{ij} \leq \gamma_{ii}$.  We say that a symmetric matrix $\Delta = [ \delta_{ij} ]$ is a {\em dissimilarity matrix}\/ if and only if each $\delta_{ij} \geq 0$ and each $\delta_{ii}=0$.  Dissimilarities are conventionally modeled with distances, whereas similarities are often modeled with inner products.  The latter interpretation is the basis for kernel methods.  We restrict attention to kernel methods because (1) most of the out-of-sample embedding methods proposed in the literature are kernel methods, and (2) it is in the context of kernel methods that the distinction between projection and restricted reconstruction is most readily understood.

Suppose that
$\gamma : \Xi \times \Xi \rightarrow \Re$ is a similarity function, i.e.,
for any objects $\xi_1,\ldots,\xi_n \in \Xi$, the $n \times n$ matrix
$\Gamma = [ \gamma(\xi_i,\xi_j) ]$ is a similarity matrix.
We center $\gamma$ with respect to $\xi_1,\ldots,\xi_n$ by computing
\begin{equation}
\tilde{\gamma}(\mu,\nu) = \gamma (\mu,\nu) -
\frac{1}{n} \sum_{j=1}^n \gamma \left( \mu, \xi_j \right) -
\frac{1}{n} \sum_{i=1}^n \gamma \left( \xi_i, \nu \right) +
\frac{1}{n^2} \sum_{i=1}^n \sum_{j=1}^n \gamma \left( \xi_i, \xi_j \right).
\label{eq:gammatilde}
\end{equation}

First, suppose that $\gamma$ is a positive definite function, i.e.,
$\Gamma$ is positive semidefinite for any $\xi_1,\ldots,\xi_n$.
Given $\xi_1,\ldots,\xi_n$, kernel PCA replaces 
$B=\tilde{M} \tilde{M}^t$ for PCA with the kernel matrix
\begin{equation}
\label{eq:B}
B = \tilde{\Gamma} = \left[ \tilde{\gamma} \left( \xi_i,\xi_j \right) \right].
\end{equation}
The derivation of an out-of-sample projection formula for the case of kernel PCA is then identical to the case of PCA.  
To embed another $\eta \in \Xi$ in the
$d$-dimensional PC representation of $\xi_1,\ldots,\xi_n$,
first compute
\begin{equation}
\label{eq:b}
b = \left[ \begin{array}{c}
\tilde{\gamma} \left( \xi_1, \eta \right) \\ \vdots \\
\tilde{\gamma} \left( \xi_n, \eta \right)
\end{array} \right],
\end{equation}
then
\begin{equation}
\hat{y} = \Sigma_d^{-1} U_d^t b.
\label{eq:kpca.b}
\end{equation}
This version of (\ref{eq:pca.x}) appears in
\cite{williams&seeger:2001}.
Bengio et al.\ \cite{bengio&etal:2003tr,bengio&etal:2003}
derived out-of-sample extensions of various kernel methods
by applying the above construction with suitable kernel functions.

To embed out-of-sample similarity data if the similarity function $\gamma : \Xi \times \Xi \rightarrow \Re$ is not positive semidefinite, replace $B$ with  
\begin{equation}
\bar{B} = U \bar{\Lambda} U^t = 
\left[ \begin{array}{c|c} U_d & \cdot \end{array} \right]
\left[ \begin{array}{c|c} \Sigma_d^2 & 0 \\ \hline 0 & 0 \end{array} \right]
\left[ \begin{array}{c} U_d^t \\ \hline \cdot \end{array} \right],
\label{eq:Bbar}
\end{equation}
then use (\ref{eq:b}) and (\ref{eq:kpca.b}).
To embed out-of-sample dissimilarity data,
suppose that 
$\delta : \Xi \times \Xi \rightarrow \Re$ is a dissimilarity function.  Converting dissimilarities to centered similarities by 
\begin{eqnarray}
\tilde{\gamma}(\mu,\nu) & = & -\frac{1}{2} \left[ \delta^2 (\mu,\nu) -
\frac{1}{n} \sum_{j=1}^n \delta^2 \left( \mu, \xi_j \right) \right. - \nonumber \\ & &
\left. \frac{1}{n} \sum_{i=1}^n \delta^2 \left( \xi_i, \nu \right) +
\frac{1}{n^2} \sum_{i=1}^n \sum_{j=1}^n \delta^2 \left( \xi_i, \xi_j \right) \right]
\label{eq:delta.to.gamma}
\end{eqnarray}
then yields the out-of-sample method proposed in 
\cite{bengio&etal:2003tr,bengio&etal:2003}.

The representation of the original objects in $\Re^d$ can be written as the $n \times d$ configuration matrix $X = U_d \Sigma_d$.  Notice that
\[
X^tX\hat{y} = \left( U_d\Sigma_d \right)^t \left( U_d\Sigma_d \right)
\left( \Sigma_d^{-1} U_d^t b \right) =
\Sigma_d^tU_d^tU_d\Sigma_d\Sigma_d^{-1}U_d^t b =
\Sigma_d^tU_d^tb = X^tb.
\]
It follows that (\ref{eq:kpca.b}) can be rewritten as
\begin{equation}
\hat{y} = \left( X^tX \right)^{-1} X^t b,
\label{eq:ols}
\end{equation}
the out-of-sample embedding formula proposed by
Anderson and Robinson \cite{anderson&robinson:2003}.

Finally, consider Landmark MDS, 
proposed in \cite{desilva&tenenbaum:2003}
by de Silva and Tenenbaum, who subsequently elaborated in 
\cite{desilva&tenenbaum:2004}.
The basic idea of Landmark MDS is to embed a large set of points by first embedding a small subset of ``landmark'' points, then positioning each additional point in relation to the landmark points.  The idea is an old one, first used by 
Kruskal and Hart \cite{kruskal&hart:1966}.  The problem of positioning each additional point is the problem of out-of-sample embedding.

Unlike Kruskal and Hart,
de Silva and Tenenbaum were seeking an alternative to applying CMDS to an entire set of points.\footnote{Isomap \cite{isomap:2000} uses CMDS to embed shortest path distances.  Isomap's embedding step is a computational bottleneck, which de Silva and Tenenbaum sought to mitigate.}
Landmark MDS embeds the landmark points by CMDS, then embeds the remaining points
by ``distance-based triangulation.''  The resulting out-of-sample embedding formula
\cite[Equation (3)]{desilva&tenenbaum:2004}
is
\begin{equation}
\label{eq:tri.x}
y = -\frac{1}{2} X^\dagger \left[ a_2 - \frac{1}{n} \Delta_2 e \right],
\end{equation}
where $\Delta_2 = [ \delta( \xi_i,\xi_j )^2 ]$ denotes the $n \times n$ matrix of pairwise dissimilarities of the original $n$ objects,
$a_2 = [ \delta( \xi,\xi_i )^2 ] \in \Re^n$ denotes the vector of squared dissimilarities of the new object from the original $n$ objects, $e=(1,\ldots,1) \in \Re^n$,
and $X^\dagger$ is the pseudoinverse of $X = U_d \Sigma_d$, i.e.,
$X^\dagger = \Sigma_d^{-1} U_d^t$.
Some manipulation reveals that
\[
-\frac{1}{2} \left[ a_2 - \frac{1}{n} \Delta_2 e \right] = b + \alpha e;
\]
hence, because $X$ is centered,
\[
X^tX y = \left( U_d\Sigma_d \right)^t U_d \Sigma_d \Sigma_d^{-1} U_d^t
(b + \alpha e) = X^tb + \alpha X^t e = X^tb,
\]
which demonstrates that (\ref{eq:tri.x}) is (\ref{eq:ols}).

Each of the above out-of-sample embedding techniques is precisely analogous to the out-of-sample projection formula for PCA.  The difference lies in the data.  PCA begins with feature vectors in Euclidean space, whereas kernel PCA constructs feature vectors from proximity data, but the projections into the original representation space are identical.

\section{Restricted Reconstruction with Proximity Data}
\label{rr}

As noted in Example~\ref{ex:3ptOut}, the general approach described in Figure~\ref{fig:strategies} requires us to identify the optimization problem solved by reconstruction and add to it constraints that fix the original configuration.  For that example, we identified PCA as finding the $d$-dimensional hyperplane that best approximates a set of feature vectors.  This characterization is both familiar and intuitive, but the modified optimization problem that results when we fix the original configuration is cumbersome and it is not obvious how to extend it to proximity data.  Instead, we turn to another well-known optimality property of PCA.

\begin{theorem}
Let $M$ denote an $n \times q$ data matrix and let
\[
\tilde{M} = U \left[ \begin{array}{c|c}
\Sigma & 0 \\ \hline 0 & 0
\end{array} \right] V^t,
\]
denote the corresponding centered data matrix with pairwise inner products $B = \tilde{M} \tilde{M}^t$.
The $d$-dimensional principal component representation of $M$
is the $n \times d$ configuration matrix $U_d \Sigma_d$, 
with pairwise inner products $\bar{B} = U_d \Sigma_d^2 U_d^t$.
Let $Z$ denote any $n \times d$ configuration matrix,
with pairwise inner products $ZZ^t$.  Then
\[
\left\| \bar{B}-B \right\|_F^2 \leq \left\| ZZ^t-B \right\|_F^2.
\]
\label{thm:PCAip}
\end{theorem}

Now suppose that $\xi_1,\ldots,\xi_n,\eta \in \Xi$ and that
$\gamma : \Xi \times \Xi \rightarrow \Re$ is a similarity function.
Write the centered similarities of $\xi_1,\ldots,\xi_n,\eta$ as
\begin{equation}
\label{eq:B+}
B^+ = \left[ \begin{array}{cc} B & b \\ b^t & \beta \end{array} \right],
\end{equation}
where $B$ is defined by (\ref{eq:B}), $b$ is defined by (\ref{eq:b}), and $\beta=\tilde{\gamma}(\eta,\eta)$ is defined using (\ref{eq:gammatilde}).
Embedding $B^+$ in $\Re^d$ entails approximating $\bar{B}^+$ with an inner product matrix of rank $d$.  Restricted reconstruction fixes the positions of $x_1,\ldots,x_n \in \Re^d$ obtained by embedding $\bar{B}$,
resulting in the nonlinear optimization problem
\begin{equation}
\label{pr:out1}
\min_{y \in \Re^d} \; \left\| 
\left[
\begin{array}{c}
x_1^t \\ \vdots \\ x_n^t \\ y^t
\end{array}
\right]
\left[
\begin{array}{cccc}
x_1 & \cdots & x_n & y
\end{array}
\right] - \bar{B}^+
\right\|_F^2 =
\min_{y \in \Re^d} \; 
2 \left\|  Xy - b \right\|_2^2 + \left( y^ty-\beta \right)^2,
\end{equation}
where $X$ is the $n \times d$ configuration matrix corresponding to $x_1,\ldots,x_n$.
We seek global solutions of (\ref{pr:out1}).

Notice that, if the quartic term $(y^ty-\beta)^2$ is dropped from the polynomial objective function in (\ref{pr:out1}), then what remains is a linear least squares problem with normal equations $X^tXy=X^tb$.
Assuming that $X$ has full rank
(otherwise, a smaller $d$ will suffice),
$X^tX$ is invertible and
the unique solution of the normal equations is (\ref{eq:ols}).
Thus, the distinction between the projection and restricted reconstruction approaches to out-of-sample embedding is entirely due to the inclusion of the quartic term in (\ref{pr:out1}), the error that results from approximating $\beta$ with $y^ty$.
Example~\ref{ex:out1} below demonstrates that including this term can have a profound effect,
i.e., that the methods of projection and restricted reconstruction may yield very different embeddings.

To embed out-of-sample dissimilarity data,
suppose $\delta : \Xi \times \Xi \rightarrow \Re$ is a dissimilarity function.  Convert dissimilarities to centered similarities by (\ref{eq:delta.to.gamma}), then proceed as above.  This is the out-of-sample embedding methodology proposed in \cite{mwt:out}.  The calculations are summarized in Figure~\ref{fig:cmds.out1}.  Notice that the use of Gower's 
\cite{gower:1982,gower:1985}
\begin{equation}
\label{eq:tau.w}
\tau_w \left( A_2 \right) =
-\frac{1}{2} \left( I-ew^t \right) A_2 \left( I-we^t \right)
\end{equation}
with $w=(1,\ldots,1,0)/n \in \Re^{n+1}$
maintains the $x_1,\ldots,x_n \in \Re^d$ constructed from the original dissimilarities by CMDS.

\begin{figure}[tb]
\begin{center}
\fbox{ 
\begin{minipage}{5in}
\vspace{1em}
\begin{enumerate}

\item Let $\Delta_2 = [ \delta_{ij}^2 ]$   
denote the squared dissimilarities of objects $\xi_1,\ldots,\xi_n$.
Embed $\Delta_2$ by CMDS, i.e.,
set $e=(1,\ldots,1) \in \Re^n$ and $P=I-ee^t/n$, then compute  
$B = -(1/2) P \Delta_2 P = U \Lambda U^t$,
\begin{eqnarray*}
\bar{B} = U \bar{\Lambda} U^t = 
\left[ \begin{array}{c|c} U_d & \cdot \end{array} \right]
\left[ \begin{array}{c|c} \Sigma_d^2 & 0 \\ \hline 0 & 0 \end{array} \right]
\left[ \begin{array}{c} U_d^t \\ \hline \cdot \end{array} \right],
 & \mbox{ and } &
X = U_d\Sigma_d.
\end{eqnarray*}

\item Let $a_2 \in \Re^n$ denote the squared dissimilarities of object $\eta$ from objects $\xi_1,\ldots,\xi_n$.  Set $w=(1,...,1,0)/n \in \Re^{n+1}$ and
\begin{eqnarray*}
A_2 = \left[
\begin{array}{cc}
\Delta_2 & a_2 \\
a_2^t & 0
\end{array}
\right],
 & \mbox{ then compute } &
\tau_w \left( A_2 \right) =
\left[  
\begin{array}{cc}
B & b \\
b^t & \beta 
\end{array} 
\right],
\end{eqnarray*}
where $\tau_w$ is defined by (\ref{eq:tau.w}).

\item Solve the nonlinear optimization problem
\[
\min_{y \in \Re^d} \;  
2 \left\| Xy-b \right\|_2^2 + \left( y^ty-\beta \right)^2
\]
to obtain $y_*$.

\end{enumerate}
\vspace{1em}
\end{minipage}
}
\end{center}
\caption{Out-of-sample embedding by restricted reconstruction for CMDS.  The original objects $\xi_1,\ldots,\xi_n$ are embedded as
$x_1,\ldots,x_n \in \Re^d$.  The out-of-sample object $\eta$ is then embedded as $y_* \in \Re^d$.}
\label{fig:cmds.out1}
\end{figure}

\begin{example}
\label{ex:out1}
Suppose that objects $\xi_1,\ldots,\xi_4$ have pairwise squared dissimilarities
\[
\Delta_2 = \left[ \delta_{ij}^2 \right] =
\left[ \begin{array}{rrrr}
0 & 100 & 45 & 45 \\
100 & 0 & 45 & 45 \\
45 & 45 & 0 & 64 \\
45 & 45 & 64 & 0
\end{array} \right].
\]
We embed $\Delta_2$ in $\Re^2$ by CMDS, obtaining the configuration matrix
\[
X = U_2 \Sigma_2 = \left[ \begin{array}{rr}
5 & 0 \\ -5 & 0 \\ 0 & 4 \\ 0 & -4 
\end{array} \right].
\]
Now suppose that another object, $\eta$, has pairwise squared dissimilarities of
\[
a_2^t = \left[ \begin{array}{rrrr}
386 & 386 & 457 & 457 
\end{array} \right]
\]
from objects $\xi_1,\ldots,\xi_4$,
from which we obtain $b = \vec{0}$ and $\beta=400$.
The objective function for restricted reconstruction,
\[
2 \left\| Xy - \vec{0} \right\|_2^2 + \left( y^ty-400 \right)^2,
\]
has two global minimizers, at $y_*=(0,\pm\sqrt{368} \doteq \pm 19.2)$.
In contrast, projection produces
$\hat{y} = \Sigma_d^{-1}U_d^t \vec{0} = (0,0)$.  The profound difference
in solutions warrants careful examination.

The configuration $x_1,x_2,x_3,x_4,y_*$ is the optimal
representation of $\xi_1,\xi_2,\xi_3,\xi_4,\eta$ in $\Re^2$,
subject to the restriction that
$\xi_1,\xi_2,\xi_3,\xi_4$ must be represented by $x_1,x_2,x_3,x_4$.
It is evident from $a_2$ that $\eta$ is more dissimilar from any original object
than any original object is from another.  This feature is preserved by
$x_1,x_2,x_3,x_4,y_*$, whereas $\hat{y}$ represents $\eta$ as the centroid of the original objects.   

Notice that $\Delta_2$ can be embedded without error in $\Re^3$, e.g., as
\[
\left[ \begin{array}{rrr}
5 & 0 & 1 \\ -5 & 0 & 1 \\ 0 & 4 & -1 \\ 0 & -4 & -1 
\end{array} \right].
\]
This is the principal component representation of $\xi_1,\xi_2,\xi_3,\xi_4$,
so the $2$-dimensional PC representation is just $X$. 
One can then embed $\eta$ in $\Re^3$ without error as
$(0,0,20)$ and observe that projecting $(0,0,20)$ into
the span of the first two principal components of $x_1,x_2,x_3,x_4$ results in
$\hat{y}=\vec{0}$.  
\end{example}

We emphasize that, unlike projection, restricted reconstruction depends on the method of construction.  Although our present concern is restricted reconstruction with kernel methods, it is instructive to consider briefly how restricted reconstruction might be formulated for another embedding technique.
Let $\delta_{ij} = \delta(\xi_i,\xi_j)$ denote the original pairwise dissimilarities and suppose that the embedding $x_1,\ldots,x_n \in \Re^d$ was obtained by minimizing Kruskal's \cite{kruskal:1964a} {\em raw stress criterion},
\[
\sigma \left( x_1,\ldots,x_n \right) =
\sum_{i,j=1}^n \left[ \left\| x_i-x_j \right\|_2 - \delta_{ij} \right]^2.
\]
To then embed $\delta_i = \delta(\xi_i,\eta)$ by restricted reconstruction, one would hold $x_1,\ldots,x_n$ fixed and choose $y \in \Re^d$ by minimizing
\[
\sigma_{\mbox{\tiny out}} (y) = \sum_{i=1}^n \left[ \left\| y-x_i \right\|_2 - \delta_i \right]^2.
\]

\section{Computing Restricted Reconstructions}
\label{compute}

Solving Problem (\ref{pr:out1}) requires finding a global minimizer of a quartic polynomial.  Finding local minimizers of quartic polynomials is easy; the potential difficulty is finding a global minimizer.  Fortunately, Problem (\ref{pr:out1}) can be reduced to a nonlinear optimization problem with a single decision variable that can then be solved by unidimensional search.  Toward that end,
define $f,g: \Re^d \rightarrow \Re$ by
\begin{eqnarray*}
f(y) = 2 \left\| b-Xy \right\|_2^2 & \mbox{and} &
g(y) = \left( \beta - y^ty \right)^2,
\end{eqnarray*}
so that Problem (\ref{pr:out1}) is the problem of minimizing $f+g$.
Let $y_* \in \Re^d$ denote a global minimizer of $f+g$ and let $r^2 = y_*^t y_*$. 
Because $g(y_1) = g(y_2)$ if $y_1^ty_1=y_2^ty_2$,
$y_*$ is also a global solution of the constrained optimization problem
\begin{eqnarray}
\mbox{minimize} & & f(y) \nonumber \\
\mbox{subject to} & & y^ty=r^2.
\label{pr:constrained}
\end{eqnarray}
Hence, we can seek $y_*$ by solving Problem (\ref{pr:constrained}) with various values of $r^2$.

We might proceed directly, by finding roots of the constrained stationary (Lagrangian) equation, but it is also instructive to
note that Problem (\ref{pr:constrained}) is a linear least squares problem with an equality constraint on the norm of the solution.  It is therefore a special case of minimizing a convex quadratic objective function subject to a spherical equality constraint, which is both the trust-region subproblem in numerical optimization and the problem that defines ridge analysis in response surface methodology.  See \cite{trosset:trust} for discussion of their equivalence.

Notice that, for each value of $r^2$, Problem (\ref{pr:constrained}) is equivalent to the regularized linear least squares problem of minimizing $f(y) + \lambda y^ty$ for some value of $\lambda$.  If $X^tX + \lambda I$ is invertible, then $\hat{y}(\lambda)$, the unique solution of the regularized problem, is obtained by solving the $d \times d$ linear system
\[
\left( X^tX + \lambda I \right)y = X^t b.
\]
Hence, we can also seek $y_*$ by minimizing
\[
\phi(\lambda) = f \left( \hat{y}(\lambda) \right) +
g \left( \hat{y}(\lambda) \right)  = 
2 \left\| b-X \hat{y}(\lambda) \right\|_2^2 +
\left[ \beta - \hat{y}(\lambda)^t \hat{y}(\lambda) \right]^2.
\]
Unlike standard examples of regularized linear least squares, however, we do not restrict the sign of $\lambda$.

Let $\hat{y}$ denote the unconstrained minimizer of $f$ and set $\hat{r} = \| \hat{y} \|_2$.  It will typically be the case that the $d$ columns of $X$ are linearly independent, in which case the $d \times d$ matrix $X^tX$ will be invertible, $\hat{y} = (X^tX)^{-1}X^tb$, and
\[
\hat{r}^2 = \hat{y}^t \hat{y} = b^tX \left( X^tX \right)^{-2} X^tb =
b^t U_1 \Sigma^{-2} U_1^tb,
\]
where $X = U_1 \Sigma  V^t$
is the thin singular value decomposition of $X$.
Notice that setting $\lambda=0$ corresponds to setting $r^2 = \hat{r}^2$ in Problem~(\ref{pr:constrained}), and therefore $\hat{y}(0)=\hat{y}$.

Let $\lambda_*$ denote the global minimizer of $\phi$, so that $\hat{y}(\lambda_*) = y_*$.  The interval of $\lambda$ that ranges from $\lambda=0$ to $\lambda=\lambda_*$ parametrizes an arc of possible embeddings from $\hat{y}$, the projection embedding, to $y_*$, the restricted reconstruction embedding.  As in ridge analysis, the entire arc may be of interest.

It is easily calculated that
\[
\beta = \delta_\cdot^2 - \delta_{\cdot \cdot}^2/2,
\]
where $\delta_\cdot^2$ is the average squared dissimilarity in $\delta_2$ and $\delta_{\cdot \cdot}^2$ is the average squared dissimilarity in $\Delta_2$.
If $\beta < \hat{r}^2$, then $\| y_* \|_2^2 < \hat{r}^2$ and $\lambda_*>0$, the case of traditional regularized linear least squares.  In this case, $X^tX + \lambda I$ is invertible for each $\lambda \in [0,\infty)$ and numerical optimization of $\phi$ by unidimensional search is numerically stable.  Furthermore, the entire arc of $\hat{y}(\lambda)$ for $\lambda \in [0,\lambda_*]$ is easily computed.

If $\beta > \hat{r}^2$, then $\| y_* \|_2^2 > \hat{r}^2$ and $\lambda_*<0$.  In this case, $X^tX + \lambda I$ is singular if (and only if) $\lambda$ is an eigenvalue of $-X^tX$ and evaluation of $\phi$ is ill-conditioned near such values.  Fortunately, $X^tX$ has at most $d$ distinct eigenvalues and $d$ is typically small.  One can use the eigenvalues of $-X^tX$ to partition $(-\infty,0]$ into at most $d+1$ subintervals.  In our experience, it is usually easy to identify the subinterval within which $\lambda_*$ lies, then conduct a unidimensional search within that subinterval to identify $\lambda_*$.  Computing the entire arc of $\hat{y}(\lambda)$ for $\lambda \in [\lambda_*,0]$ may be more challenging, as $[\lambda_*,0]$ may contain eigenvalues of $-X^tX$.  In such instances, it makes sense to interpolate $\hat{y}(\lambda)$ in the neighborhoods of offending $\lambda$.

\section{Discussion}
\label{disc}

We now return to the differences between the methods proposed in \cite{bengio&etal:2003tr,bengio&etal:2003} and in \cite{mwt:out} that motivated this investigation.  Both methods can be understood in terms of the matrix (\ref{eq:Bbar+}), in which $\bar{B}$ is held fixed while $b$ and/or $\beta$ are approximated.  The method of 
\cite{bengio&etal:2003tr,bengio&etal:2003} 
approximates $b$ without regard to $\beta$,
whereas the method of \cite{mwt:out} approximates both $b$ and $\beta$.
Bengio et al.\ \cite[Proposition 3]{bengio&etal:2003} 
clearly understood that this was the case, as they explicitly rejected the possibility of minimizing (\ref{pr:out1}):
\begin{quote}
  ``Note that by doing so, we do not seek to approximate
  $\beta = \tilde{\gamma}(\eta,\eta)$.  Future work should investigate
  embeddings which minimize the empirical reconstruction error of
  $\tilde{\gamma}$ but ignore the diagonal contributions.''\footnote{We have quoted
  Bengio et al.\ \cite[Section 4]{bengio&etal:2003} verbatim, 
  but substituted our notation for theirs.}
\end{quote}
Tang and Trosset \cite{tang&trosset:out2} retorted that ``CMDS does not ignore diagonal contributions and neither should out-of-sample extensions of CMDS---unless one seeks a less expensive approximate solution.''  Each group of researchers believed their approach to be correct.
Years later, we have come to embrace a more enlightened perspective.
We now understand that each method can be derived from a general principle.  The question of whether or not to approximate $\beta$ is ultimately a question of whether to deploy projection or restricted reconstruction.  We do not believe that this question has a single answer; accordingly, we conclude by discussing some circumstances that may warrant one or the other.

Consider the method of Landmark MDS, which efficiently constructs a representation of $n+k$ objects by first applying CMDS to a small set of $n$ landmark objects, then embedding each of the remaining $k$ objects in that representation.  The resulting representation might be an end in itself, or it might be an intermediate step in a more elaborate embedding procedure.  (For example, iterative methods for minimizing the raw stress criterion require an initial configuration from which optimization is to commence.)  The point of Landmark MDS is that the landmarks determine the representation space, so that all that needs to be done to embed the remaining objects is to project each in turn into a fixed representation space.  Restricted reconstruction would not make sense in this context.

If $\hat{y}$ (the solution obtained by projection) and $y_*$ (the solution obtained by restricted reconstruction) differ substantially, then it must be that $\eta$ substantially differs from $\xi,\ldots,\xi_n$ in ways that are not captured by the original representation space.  If such a difference does not matter to the data analyst, then the original representation space is adequate and representing $\eta$ by $\hat{y}$ is appropriate.  If such a difference does matter, then the original representation space is not adequate and representing $\eta$ by $y_*$ is appropriate.

Finally, consider the case of two new objects, $\eta_1$ and $\eta_2$, that are known to differ substantially from each other.  Does this difference matter?  For example, suppose that we modify Example~\ref{ex:3ptOut} so that $\eta_1=(0,9)$ and $\eta_2=(0,-9)$.  Projecting $\eta_i$ into the horizontal axis of $\Re^2$ results in $\hat{y_i} = \hat{\eta_i} = 0$ for each $i$, despite the fact that $\| \eta_1-\eta_2 \|_2 =18$ is by far the largest of the pairwise distances between the four feature vectors in $\Re^2$.  Projection takes no notice of this fact.  In contrast, restricted reconstruction readily accommodates proximity information about multiple new points.

Let $\xi_1,\ldots,\xi_n \in \Xi$ denote the original objects and $\eta_1,\ldots,\eta_k \in \Xi$ denote the new objects.  The restricted reconstruction procedure summarized in Figure~\ref{fig:cmds.out1} is then modified as follows, as in \cite{mwt:out}.
\begin{quote}
Let $a_2$ denote the $n \times k$ matrix of squared dissimilarities between $\xi_i$ and $\eta_j$, and let $\alpha_2$ denote the $k \times k$ matrix of squared dissimilarities between $\eta_i$ and $\eta_j$.
Set $w=(1,...,1,0,\ldots,0) \in \Re^{n+k}$ and
\begin{eqnarray*}
A_2 = \left[
\begin{array}{cc}
\Delta_2 & a_2 \\
a_2^t & \alpha_2
\end{array}
\right], & \mbox{ then minimize } &  
2 \left\| XY^t-b \right\|_F^2 + \left\| YY^t-\beta \right\|_F^2
\end{eqnarray*}
over $Y \in \Re^{k \times d}$ to obtain the $k \times d$ configuration matrix
\[
Y_* = \left[ \begin{array}{c} y_{1*}^t \\ \vdots \\ y_{k*}^t \end{array} \right].
\]
\end{quote}
In this setting, the choice that confronts the data analyst is stark: either (1) discard all information about how the new objects relate to each and individually project each new object into the original representation space, or (2) construct a new representation space that preserves the configuration of the original objects and positions the new objects so that {\em all}\/ of the new proximity information is approximated.  In the applications that we have encountered, e.g., in \cite{mwt:semi}, we have preferred the latter construction.

\section*{Acknowledgments}
This work was partially supported by the Naval Engineering Education Consortium (NEEC), Office of Naval Research (ONR) Award Number N00174-19-1-0011; ONR Award Number N00024-22-D-6404 (via Johns Hopkins University APL); and ONR Award Number N00014-24-1-2278 (Science of Autonomy).
The first author benefitted from discussions with Julia Fukuyama.

\bibliography{$HOME/lib/tex/stat,$HOME/lib/tex/mds,$HOME/lib/tex/math,$HOME/lib/tex/mwt,$HOME/lib/tex/cep,$HOME/lib/tex/bio,$HOME/lib/tex/proximity,$HOME/lib/tex/na1,$HOME/lib/tex/na2}

\end{document}